\documentclass[runningheads]{llncs}
\usepackage[nocompress]{cite}
\usepackage{amsmath,amsfonts}
\usepackage{algorithmic}
\usepackage{algorithm}
\usepackage{array}
\usepackage[caption=false,font=normalsize,labelfont=sf,textfont=sf]{subfig}
\usepackage{textcomp}
\usepackage{stfloats}
\usepackage{url}
\usepackage{verbatim}
\usepackage{graphicx}
\usepackage{cite}
\usepackage{balance}
\usepackage{soul}
\usepackage{multirow}
\usepackage{perpage}
\usepackage[table,xcdraw]{xcolor}
\usepackage[skip=0pt]{caption}
\usepackage[titletoc]{appendix}
\usepackage[misc,geometry]{ifsym}
\usepackage{hyperref}
\hypersetup{pdfborder={0 0 0}}
\usepackage{indentfirst}

\begin{document}
\title{Balanced Graph Structure Information for Brain Disease Detection}

% \author{Falih Gozi Febrinanto\inst{1} \orcidID{0000-0002-7932-4571} \and
% Mujie Liu\inst{1} \orcidID{0009-0002-0879-7168} \and
% Feng Xia\inst{2} \orcidID{0000-0002-8324-1859}
% }

\author{Falih Gozi Febrinanto\inst{1} \and
Mujie Liu\inst{1} \and
Feng Xia\inst{2}
}

\authorrunning{Febrinanto et al.}

\institute{Institute of Innovation, Science, and Sustainability, Federation University Australia, Ballarat, Australia \\
\email{\{f.febrinanto, mujie.liu\}@federation.edu.au}
\and
School of Computing Technologies, RMIT University, Melbourne, Australia \\
\email{f.xia@ieee.org}
}

\maketitle
\begin{abstract}
Analyzing connections between brain regions of interest (ROI) is vital to detect neurological disorders such as autism or schizophrenia. Recent advancements employ graph neural networks (GNNs) to utilize graph structures in brains, improving detection performances. Current methods use correlation measures between ROI's blood-oxygen-level-dependent (BOLD) signals to generate the graph structure. Other methods use the training samples to learn the optimal graph structure through end-to-end learning. However, implementing those methods independently leads to some issues with noisy data for the correlation graphs and overfitting problems for the optimal graph. In this work, we proposed \textbf{Bargrain} (balanced graph structure for brains), which models two graph structures: filtered correlation matrix and optimal sample graph using graph convolution networks (GCNs). This approach aims to get advantages from both graphs and address the limitations of only relying on a single type of structure. Based on our extensive experiment, Bargrain outperforms state-of-the-art methods in classification tasks on brain disease datasets, as measured by average F1 scores. 

\keywords{Brain Network \and Classification \and Graph Learning  \and Graph Neural Networks \and Disease Detection}
\end{abstract}

\section{Introduction} 
Resting-state functional magnetic resonance imaging (Rest fMRI) analysis is vital for detecting brain diseases such as autism or schizophrenia in individuals~\cite{kan2022brain, cui2022braingb, kan2022fbnetgen}. FMRI data measures changes in blood oxygen level-dependent signals (BOLD) over a specific time, providing essential information on brain activity~\cite{hutchison2013dynamic_issues,elgazzar2022benchmarking}. These BOLD signals represent variations in blood oxygen levels in different regions of interest (ROI), which are generated using several atlas techniques to divide the brain into distinct regions~\cite{li2021braingnn, elgazzar2022benchmarking}. Researchers have a huge consensus on analyzing interactions between regions in brain networks as a key to a better diagnosis for detecting brain diseases. 

Graph neural networks (GNN) have shown promising results in improving performance to predict diseases in the brain networks~\cite{kan2022brain,dadi2019benchmarking}. However, there is still a challenge in defining the appropriate structure of graphs in brain networks. Some methods~\cite{elgazzar2022benchmarking, li2021braingnn} adopt a 
correlation matrix to generate graphs that calculate the similarity between series of BOLD signals across all brain regions. This technique has a good aspect in incorporating biological insight or actual domain knowledge of brain structure. However, this technique potentially leads to inaccurate correlations matrix due to some noise caused by scanner drift or physiological noise that arises from cardiac pulsation, shifts caused by the body's motion~\cite{hutchison2013dynamic_issues}. Other methods~\cite{kan2022fbnetgen, kazi2022differentiable, kan2022brain} disregard the domain knowledge structure and instead utilize a learnable graph structure to search for optimal structures over variations in sample data through an end-to-end learning process, making it resistant to noise. However, these methods fail to enhance the interpretability of biological insights into graph structures and lack generalization to unseen data samples, making them prone to overfitting to training samples~\cite{chen2023balanced}.

\setcounter{footnote}{0}
To address those problems, we propose a framework that utilizes balanced structure graphs for brain disease classification, called \textbf{Bargrain} (balanced graph structure for brains)~\footnote{The implementation of Bargrain: \textbf{\url{https://github.com/falihgoz/Bargrain}}}. It combines predefined signal correlation and learnable methods to generate the graph, aiming for both brain network information advantages. Our model applies a \textit{filtered correlation matrix graph} based on signal similarities and an \textit{optimal sampling graph} based on the Gumbel reparameterization trick~\cite{jang2017categorical_gumbel}. This helps us prevent the noise issues in solely using domain knowledge structure and overfitting problems from just relying on learning optimal structure. We leverage effective node features derived from the ROIs' correlation matrix and employ graph convolutional networks (GCN)~\cite{welling2016semi_gcn} for modeling the graph structures. We also use a graph readout function based on a CONCAT pooling operator since the disease prediction is a graph-level task. We summarize the main contribution of this paper as follows:

\begin{itemize}
  \item We propose Bargain, a brain disease detection framework that utilizes a balanced graph structure by merging two valuable insights: actual domain knowledge structure and optimal structure of brain networks.
  \item We conduct an extensive experiment on real-world brain datasets. Our experimental results demonstrate that our method outperforms state-of-the-art models in classifying brain diseases.
  \item We systematically review how the two graph structures differ in network visualizations and node degree distributions, which enhances brain disease detection by using their complementary information.
\end{itemize}

\section{Related Work}
\subsection{GNNs for Brain Disease Detection} Graph Neural Networks (GNNs) have demonstrated impressive performance in tackling various detection tasks involving graph data, including their application in healthcare for detecting brain diseases~\cite{ren2023graph}. To represent the graph structure in the brain, the correlation matrix is a common method used to define the connectivity between ROIs. For example, GDC-GCN~\cite{elgazzar2022benchmarking} follows the correlation calculation to create a graph structure and uses a graph diffusion technique to reduce noisy structures. BrainGNN~\cite{li2021braingnn} develops ROI-aware GNNs to utilize a special pooling strategy to select important nodes. On the other hand, instead of using predefined graphs, FBNetGen~\cite{kan2022fbnetgen} explores a learnable graph to create brain networks and investigate some downstream tasks. Besides, DGM~\cite{kazi2022differentiable} designs a latent-graph learning block to build a probabilistic graph. Inspired by the advantages of graph transformers, BrainNETTF~\cite{kan2022brain} capitalizes on the distinctive characteristics of brain network data.

\subsection{Learnable Graph Generations} Learnable graph generation strategy aims to create optimal graphs through an end-to-end learning process. In practical scenarios, the graph structure is not always available or might be incomplete~\cite{shang2020discrete_gts, febrinanto2023efficient}. On top of that, even if a predefined graph structure exists, it might not provide the necessary information or align with the requirements of downstream tasks. Precisely, a learnable graph generation adjusts its structure based on the available data of all nodes, capturing intricate relationships that predefined graphs could overlook~\cite{chen2023balanced}. This adaptability improves the model's capability to unveil subtle, data-specific connections from provided samples, which can help mitigate noise in the predefined graph structure.

Some techniques use a learnable node representation to calculate cosine similarity between those representations and optimize them during the learning process~\cite{kan2022fbnetgen,kazi2022differentiable}. However, most of them apply the top-$k$ closest relations to maintain graph sparsity, which impedes the designs of model flexibility and potentially eliminates some vital information. Therefore, considering the inequality of the information problem~\cite{xia2022cengcn} in the message-passing process, the optimal sampling techniques~\cite{shang2020discrete_gts} based on categorical reparameterization trick\cite{jang2017categorical_gumbel}, which enables the approximation of samples from a categorical distribution, is introduced as an alternative to improve the flexibility by not selecting the top-$k$ nearest nodes.

\subsection{Framework Overview}
\begin{figure*}[!t]
  \begin{center}
    \includegraphics[width=.92\textwidth]{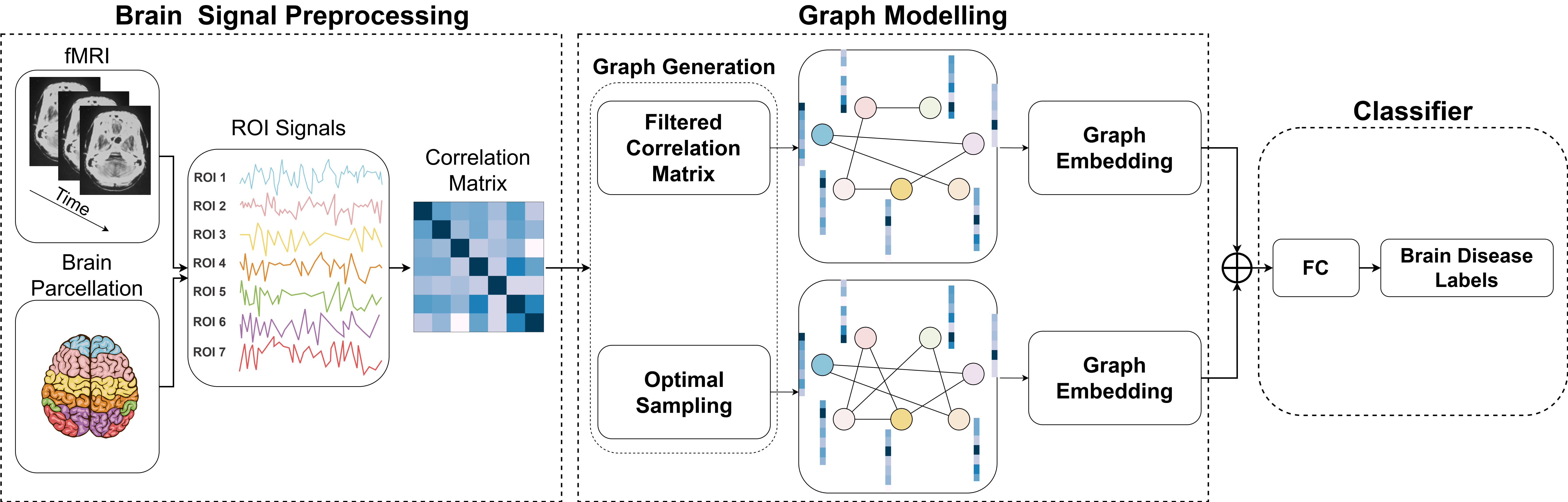}
  \end{center}
  \caption{The proposed framework. \textit{Bargrain} starts with \textit{Brain Signal Preprocessing} to develop a correlation matrix based on ROI signals. \textit{Graph Modeling} processes two graph structures to balance information from domain knowledge and optimal structures. \textit{Classifier} maps the combined knowledge for brain disease detection.}
  \label{img:fram}
\end{figure*}

Bargrain's overall framework is shown in Figure~\ref{img:fram}. Based on our proposed method, there are 3 modules to classify brain disease.

\begin{itemize}
  \item \textbf{Brain Signal Preprocessing}. We calculate the correlation matrix among the brain's regions of interest (ROIs). The developed correlation matrix will serve as both node features and the foundation for generating graph structure in the subsequent phase.
  \item \textbf{Graph Modeling}. Our objective is to balance the information employing both graph structures. We perform two graph structure generations: filtered correlation matrix graph and optimal sampling graph. The next step is to learn the spatial information for each graph structure using a graph convolution network (GCN). The resulting two graph-level representations, after the pooling process, will be used in the next module.
  \item \textbf{Classifier}. This step aims to combine two graph-level representations and map the combined knowledge to perform brain disease classification.
\end{itemize}

\section{Proposed Framework}

\subsection{Brain Signal Preprocessing}
Each brain's ROI produces signal series, and overall signals are represented by matrix $X \in \mathbb{R}^{N\times T}$, where $N$ is the number of nodes (ROIs), and $T$ represents a period of recording time. From these signal series, a correlation matrix $V \in \mathbb{R}^{N\times N}$ is generated to assess the similarity signals between ROIs. Moreover, the correlation values are utilized as node features in the graph, demonstrating a significant performance in detecting brain diseases~\cite{kan2022brain,elgazzar2022benchmarking}.

\subsection{Graph Modeling}
\label{sec:grapmod}
\noindent \textbf{Graph Generation}. To balance the information about the graph structure, our framework's graph generation process is divided into two: \textit{filtered correlation matrix graph} and \textit{optimal sampling graph}.

The \textit{filtered correlation matrix graph} aims to preserve the actual domain knowledge of the brain structure. The adjacency matrix of this graph type denotes as $A^{\textup{filtered}} \in \mathbb{R}^{N\times N}$. To maintain the sparsity of the graph, it is obtained by comparing the correlation matrix $V \in \mathbb{R}^{N\times N}$ with a specific threshold $c$, as follows:

\begin{equation}
  \label{eq:adj}
  A^{\textup{filtered}}_{ij} =
  \begin{cases}
    1           & \text{if $V_{ij} > c;$} \\
    0           & \text{otherwise}
  \end{cases}
\end{equation}

The \textit{optimal sampling graph} aims to find the optimal structure based on the given training samples of the brain ROIs through an end-to-end learning process. The adjacency matrix of this type of graph is denoted as $A^{\textup{optimal}} \in \mathbb{R}^{N \times N}$. The binary values in adjacency matrix $A \in \{0,1\} ^{N \times N}$ are originally non-differentiable with the typical backpropagation due to their discrete nature. To solve this problem, the Gumbel reparameterization trick~\cite{jang2017categorical_gumbel} was proposed. With that idea, we employ the Gumbel reparameterization trick to sample the optimal graph structure in the brain network. The Gumbel reparameterization trick to learn the graph structure is as follows: 

\begin{equation}
  \label{eq:gumble}
  A^{\textup{optimal}}_{ij} = \textup{sigmoid}((log(\theta_{ij}/(1 - \theta_{ij})) + (g^1_{ij} - g^2_{ij})/\tau),
\end{equation}

\noindent where $\theta_{ij}$ is learnable features vector based on node $v_i$ and $v_j$, $g^1_{ij} \textup{,} g^2_{ij} \sim \textup{Gumbel(0,1)}$ for all $i$,$j$, and $\tau$ represents temperature to control Gumbel distribution. High sigmoid probability represents a relation $A^{\textup{optimal}}_{ij} = 1$ or 0 otherwise. As mentioned above, $\theta_{ij}$ is a learnable feature vector based on two nodes $v_i$ and $v_j$. We perform a feature extraction mechanism~\cite{shang2020discrete_gts} to encode signal representation in each ROI node $v_i$ to a vector $h_i$ for each $i$. We concatenate two embedding node vectors and apply two fully connected layers so that $\theta_{ij} = FC(FC(h^i || h^j))$. 

Calculating the similarity between the brain's ROIs in the filtered correlation matrix graph generation will result in an undirected graph structure. It differs from the optimal sampling graph that results in a directed graph structure due to its sampling process.

\noindent \textbf{Graph Embedding}. 
We utilize a 2-layer graph convolutional networks (GCN)~\cite{welling2016semi_gcn} to model the spatial information in the generated brain networks. The reason for using the 2-layer GCN is to balance simplicity and expressive power to model spatial information. The 2-layer GCN model with input node feature matrix $V$ and adjacency matrix $A$ can be expressed as:

\begin{equation}
  \label{eq:gnn}
  f(V, A) = \textup{ReLU}(\widetilde{A}\,\textup{ReLU}(\widetilde{A}VW_0)W_1),
\end{equation}

\noindent where $\widetilde{A} = \hat{D}^{-\frac{1}{2}}\hat{A}\hat{D}^{-\frac{1}{2}}$ denotes the normalized adjacency matrix with $\hat{A}=A + I$ is a normalized adjacency matrix that adds self-loops to the original graph and $\hat{D}$ is the diagonal degree matrix of $\hat{A}$. $W_0 \in \mathbb{R}^{N\times H}$ denotes the weight matrix from input to the first hidden layer, $W_1 \in \mathbb{R}^{H\times F}$ represents the weight matrix from the hidden layer to the output, $f(V, A) \in \mathbb{R}^{N\times F}$ represents the output with sequence length $F$, and $\textup{ReLU}(\cdot)$ is rectified linear unit, an activation function.

The two graphs are modelled parallelly: $f(V, A^{\textup{filtered}})$ and $f(V, A^{\textup{optimal}})$. Since the brain classification task is based on graph-level classification, we need to implement a \textit{graph pooling} operator for both graph embeddings. The CONCAT pooling operator combines all node embeddings into a single vector to maintain overall information. Two vectors are generated from graph pooling implementation: $\Hat{z}^{\textup{filtered}} = \textup{POOL} (f (V, A^{\textup{filtered}}))$ and $\Hat{z}^{\textup{optimal}} = \textup{POOL} (f (V, A^{\textup{optimal}}))$. 

\subsection{Classifier}

Once we get the graph representation $\Hat{z}^{\textup{filtered}}$ and $\Hat{z}^{\textup{optimal}}$, we perform concatenation to combine the information from both spatial information as follows: $(\Hat{z}^{\textup{filtered}} || \Hat{z}^{\textup{optimal}})$. Then we apply two fully connected layers to implement binary disease classification so that is: $y = \textup{FC}(\textup{FC}(\Hat{z}^{\textup{filtered}} || \Hat{z}^{\textup{optimal}}))$. The output $y$ is mapped into $1$, indicating a specific brain disease or $0$ for a normal sample.

\section{Experiment}
In this experiment section, we aim to answer 3 research questions: \textbf{RQ1}: Does Bargrain outperform baseline methods in terms of accuracy for brain disease detection? \textbf{RQ2}: Do various components in Bargrain contribute to the overall model performance? \textbf{RQ3}: Are there underlying differences between the two kinds of graph structures employed in Bargrain model to use as complementary information?

\subsection{Datasets}
We conduct experiments with 3 open-source fMRI brain classification datasets: First, Cobre~\footnote{http://cobre.mrn.org/} records fMRI data from 72 patients with schizophrenia and 75 healthy controls. We preprocessed the Cobre dataset with 150 temporal signal steps and 96 ROI nodes. Second, ACPI~\footnote{http://fcon\_1000.projects.nitrc.org/indi/ACPI/html/} records fMRI data to classify 62 patients with marijuana consumption records and 64 with healthy controls. We preprocessed the ACPI dataset with 700 temporal signal steps and 200 ROI nodes. Third, ABIDE~\cite{di2014autism_abide} records fMRI data from 402 patients with autism and 464 healthy controls. We preprocessed the ABIDE dataset with 192 temporal signal steps and 111 ROI nodes.

\subsection{Baseline Approaches and Reproducibility}
We compare the performance of our proposed method with the latest brain disease detection frameworks such as FBNetGNN~\cite{kan2022fbnetgen}, DGM~\cite{kazi2022differentiable}, BrainNetCNN~\cite{kawahara2017brainnetcnn}, BrainNETTF~\cite{kan2022brain} and GDC-GCN~\cite{elgazzar2022benchmarking}. We used their original code implementations to carry out the experiment.

\subsection{Model Selection and Experiment Setup} 
We divide each data set into training sets (80\%) and testing sets (20\%). Furthermore, we again split the training sets into actual training sets (85\%) and validation sets (15\%) to help model selections. 

We used AMD Rayzen 7 5800H @ 3.20 GHz with NVIDIA GeForce RTX 3050 Ti GPU to run the experiments. The model was trained by Adam optimizer with a learning rate $1 \times 10^{-4}$. In the first layer GCN, we used a 256 embedding size for Cobre and ACPI, and 64 for ABIDE. In the second layer GCN, we used 256 embedding sizes for Cobre and ACPI, and 512 for ABIDE. To ensure the sparsity of the filtered correlation matrix graph, we set 0.6 as a threshold $c$. Moreover, we configured $\tau = 1$ for the temperature in the Gumbel reparameterization trick.

\subsection{RQ1. Performance Comparison}
The brain disease classification results on three datasets are presented in table~\ref{tb:results}. Our model achieved the best with an average F1-score of 0.7329 across three datasets. By incorporating both actual domain knowledge and optimal structure graph information, Bargrain is able to enhance classification performance. Compared with the methods that solely rely on domain knowledge graphs, such as GDC-GCN, and the techniques that only use optimal graph structure, such as DGM, our approach outperforms them in classifying brain disease. However, it should be noted that our balanced structure graphs do not necessarily reduce the model complexity since we combine both types of information. Furthermore, There is potential for further research to explore optimizing the efficiency of models while maintaining a balance between those two graph structures and performing incremental learning when there is a concept drift in the current knowledge~\cite{febrinanto2023graph}.
\begin{table*}[!t]
\centering
\caption{The experiment results based on F1 score, sensitivity, specificity, and area under the receiver operating characteristic curve (ROC/AUC).}
\label{tb:results}
\resizebox{1\textwidth}{!}{
\begin{tabular}{l|llll|llll|llll|l}
\hline
\multicolumn{1}{c|}{\multirow{2}{*}{\textbf{Methods}}} & \multicolumn{4}{c|}{\textbf{Cobre}} & \multicolumn{4}{c|}{\textbf{ACPI}} & \multicolumn{4}{c|}{\textbf{ABIDE}} & \multicolumn{1}{c}{\multirow{2}{*}{\textbf{\begin{tabular}[c]{@{}c@{}}Average\\ F1\end{tabular}}}} \\ \cline{2-13}
\multicolumn{1}{c|}{} & \textbf{F1} & \textbf{Sens} & \textbf{Spec} & \textbf{AUC} & \textbf{F1} & \textbf{Sens} & \textbf{Spec} & \textbf{AUC} & \textbf{F1} & \textbf{Sens} & \textbf{Spec} & \textbf{AUC} & \multicolumn{1}{c}{} \\ \hline
FBNetGNN & 0.5600 & 0.5000 & 0.7333 & 0.6167 & 0.6000 & 0.7500 & 0.3077 & 0.5288 & 0.6486 & 0.6383 & 0.6173 & 0.6278 & 0.6029 \\
DGM & 0.6400 & 0.5714 & 0.8000 & 0.6857 & 0.7200 & 0.7500 & \textbf{0.6923} & 0.7212 & 0.5631 & 0.617 & 0.3333 & 0.4752 & 0.6410 \\
BrainNetCNN & 0.6923 & 0.6429 & 0.8000 & 0.7214 & 0.5833 & 0.5833 & 0.6154 & 0.5994 & 0.7158 & 0.7234 & \textbf{0.6543} & \textbf{0.6889} & 0.6638 \\
BrainNETTF & 0.5217 & 0.4286 & 0.8000 & 0.6143 & 0.6400 & 0.6667 & 0.6154 & 0.6410 & 0.6984 & 0.7021 & 0.6420 & 0.6721 & 0.6200 \\
GDC-GCN & 0.5185 & 0.5000 & 0.6000 & 0.5500 & 0.6250 & 0.8333 & 0.2308 & 0.5321 & 0.6377 & 0.7021 & 0.4198 & 0.5609 & 0.5937 \\
\textbf{Bargrain} & \textbf{0.7407} & \textbf{0.7143} & 0.8000 & \textbf{0.7571} & \textbf{0.7407} & 0.8333 & 0.6154 & \textbf{0.7244} & \textbf{0.7172} & \textbf{0.7553} & 0.5926 & 0.674 & \textbf{0.7329} \\ \hline
\end{tabular}}
\end{table*}

\begin{table*}[!b]
\small
\centering
\caption{Ablation Studies}
\label{tb:ablation}
\resizebox{.5\textwidth}{!}{
\begin{tabular}{l|lll|l}
\hline
\multirow{2}{*}{Methods} & \multicolumn{3}{c|}{F1} & \multicolumn{1}{c}{\multirow{2}{*}{Average F1}} \\ \cline{2-4}
 & Cobre & ACPI & ABIDE & \multicolumn{1}{c}{} \\ \hline
Bargrain & \textbf{0.7407} & \textbf{0.7407} & \textbf{0.7172} & \textbf{0.7329} \\ \hline
\textit{- CorrGraph} & 0.6923 & 0.6667 & 0.6845 & 0.6812 \\
\textit{- OptimGraph} & 0.6923 & 0.6086 & 0.5028 & 0.6012 \\
\textit{- GConv} & 0.6400 & 0.6086 & 0.6588 & 0.6358
\\ \hline
\end{tabular}}
\end{table*}

\subsection{RQ2. Ablation Studies}
We evaluate the effectiveness of the components of our model. To do this, we perform ablation studies by excluding specific components of our model. There are three ablation settings. We exclude each graph structure and instead maintain only one graph structure. First, without the filtered correlation matrix graph (\textbf{\textit{-CorrGraph}}) and second, without the optimal sampling graph (\textbf{\textit{-OptimGraph}}). The last is without graph convolution (\textbf{\textit{-GConv)}}, which removes the graph embedding process.

The results of the ablation studies, presented in Table~\ref{tb:ablation}, demonstrate that the most effective components for improving brain disease detection performance within the Bargrain framework are the full framework components.

\subsection{RQ.3 Brain's Graph Structure Interpretation}
\begin{figure*}[!t]
    \centering
    \begin{tabular}{@{}c@{}}
        \includegraphics[width=0.6\textwidth]{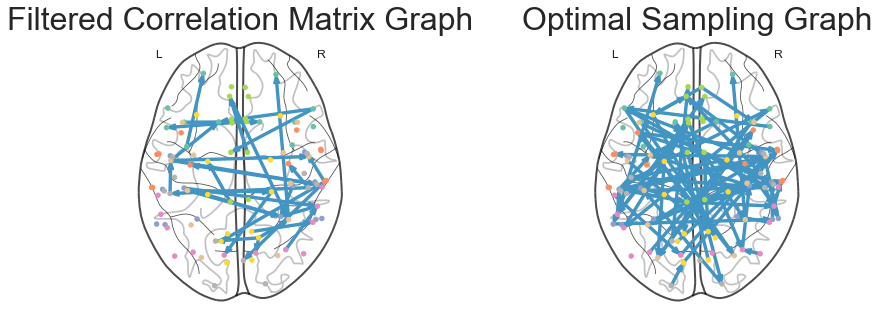} \\[\abovecaptionskip]
        \small (a) For visualization purposes, we randomly choose the top 2\% of edges to display.
        \label{subfig:adj}
    \end{tabular}
    
    \vspace{0.5em} 
    
    \begin{tabular}{@{}c@{}}
        \includegraphics[width=0.9\textwidth]{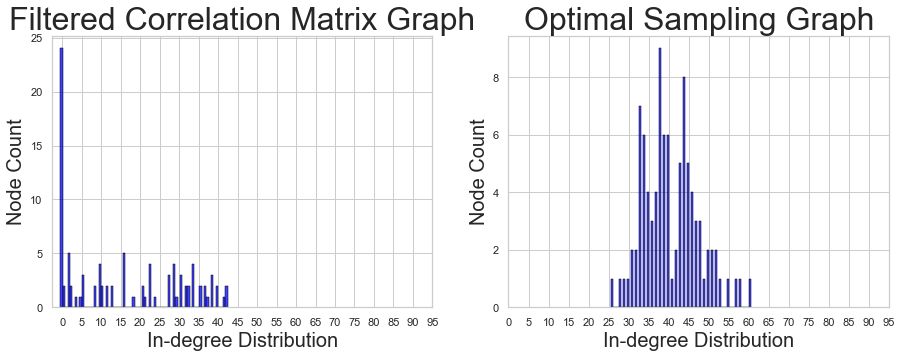} \\[\abovecaptionskip]
        \small (b) In-degree distribution of each brain graph structure.
        \label{subfig:degree}
    \end{tabular}

    \vspace{0.5em} 
    
    \caption{Balanced graph structure interpretation for a person in the Cobre dataset.}
    \label{img:adjdegree}
\end{figure*}

This section presents visualizations highlighting the characteristic differences between the two types of graph structures. In terms of their relationship characteristics (refer to Subsection~\ref{sec:grapmod}), the filtered correlation matrix exhibits an undirected structure, while the optimal sampling graph has a directed structure. Additionally, Figure~\ref{img:adjdegree}.a presents visualizations of two distinct brain graph structures from a single person in the Cobre dataset. For visualization purposes, we randomly selected the top 2\% generated edges within each brain network type. A noticeable difference emerges: the optimal sampling graph tends to have a denser graph than the filtered correlation matrix, which brings another information perspective.

Furthermore, as depicted in Figure~\ref{img:adjdegree}.b, these graphs exhibit distinct variations in their in-degree distributions, indicating the number of edges entering a specific node. The optimal sampling graph displays a bell-shaped distribution around the mid-range, with most nodes having in-degree edges ranging from 35 to 45. Notably, all nodes within the optimal sampling graph maintain connections with other nodes, unlike the filtered correlation matrix graph showing some nodes without connection.

Leveraging the diverse insights from both graphs is the primary goal of Bargrain. Thus, based on those interpretations, integrating those structures within the learning model becomes necessary to enhance prediction accuracy.

\section{Conclusion}
We proposed a brain disease detection method called \textbf{Bargrain} (balanced graph structure for brains). It employs two graph generation techniques: a filtered correlation matrix and an optimal sampling graph. Modeling those two graph representations balances the domain knowledge structure based on actual biological insight and the learnable optimal structure to prevent some noisy relations. Our method demonstrates a great performance compared to the state-of-the-art models, as shown in our extensive experiment. In our future works, we desire to implement a data-efficient approach to reduce the complexity of models for having dense relations based on high numbers of nodes in brain networks.

\bibliographystyle{splncs04}
\bibliography{ref.bib}

\end{document}